\documentclass[10pt,conference]{IEEEtran}
\IEEEoverridecommandlockouts
\usepackage{cite}
\usepackage{amsmath,amssymb,amsfonts}
\usepackage{algorithmic}
\usepackage{graphicx}
\usepackage{textcomp}
\usepackage{xcolor}
\usepackage{bbm}
\usepackage{authblk}
\usepackage{comment}
\usepackage{subcaption}  
\usepackage{booktabs}
\usepackage{xcolor}
\usepackage{url}
\usepackage{bookmark}

\usepackage{comment}
\usepackage{todonotes}
\usepackage{fancyhdr}
\def\BibTeX{{\rm B\kern-.05em{\sc i\kern-.025em b}\kern-.08em
    T\kern-.1667em\lower.7ex\hbox{E}\kern-.125emX}}

\begin{document}

\IEEEoverridecommandlockouts
\IEEEpubid{\makebox[\columnwidth]{\textbf{Preprint version. Accepted for publication at IEEE ICECCME 2025.} \hfill} \hspace{\columnsep}\makebox[\columnwidth]{}}

%\title{FedMultiEmo: Multimodality-based Real-time Emotion Recognition Using Federated Learning\\
\title{FedMultiEmo: Real-Time Emotion Recognition via Multimodal Federated Learning}

\author[1]{Baran Can Gül}
\author[2]{Suraksha Nadig}
\author[1,3]{Stefanos Tziampazis}
\author[1]{Nasser Jazdi}
\author[1]{Michael Weyrich}
%\author[1]{ }

\affil[1]{\footnotesize Institute of Industrial Automation and Software Engineering, University of Stuttgart, Germany}
\affil[ ]{\footnotesize \texttt{\{baran-can.guel, nasser.jazdi, michael.weyrich\}@ias.uni-stuttgart.de}}
\affil[2]{\footnotesize Electrical Engineering, University of Stuttgart, Germany}
\affil[ ]{\footnotesize \texttt{st184494@stud.uni-stuttgart.de}}

\affil[3]{\footnotesize Graduate School of Excellence advanced Manufacturing Engineering, University of Stuttgart, Germany}
\affil[ ]{\footnotesize \texttt{stefanos.tziampazis@gsame.uni-stuttgart.de}}

\affil[*]{\footnotesize Corresponding author. \textit{E-mail address:} baran-can.guel@ias.uni-stuttgart.de}

\maketitle

\begin{abstract}
In-vehicle emotion recognition underpins adaptive driver-assistance systems and, ultimately, occupant safety. However, practical deployment is hindered by (i) \textit{modality fragility}---poor lighting and occlusions degrade vision-based methods; (ii) \textit{physiological variability}---heart-rate and skin-conductance patterns differ across individuals; and (iii) \textit{privacy risk}---centralized training requires transmission of sensitive data.
To address these challenges, we present \textbf{FedMultiEmo}, a privacy-preserving framework that fuses two complementary modalities at the decision level: visual features extracted by a Convolutional Neural Network from facial images, and physiological cues (heart rate, electrodermal activity, and skin temperature) classified by a Random Forest. FedMultiEmo builds on three key elements: (1) a multimodal federated learning pipeline with majority-vote fusion, (2) an end-to-end edge-to-cloud prototype on Raspberry Pi clients and a Flower server, and (3) a personalized Federated Averaging scheme that weights client updates by local data volume.
Evaluated on FER2013 and a custom physiological dataset, the federated Convolutional Neural Network attains 77\% accuracy, the Random Forest 74\%, and their fusion 87\%, matching a centralized baseline while keeping all raw data local. The developed system converges in 18 rounds, with an average round time of 120~s and a per-client memory footprint below 200~MB. These results indicate that FedMultiEmo offers a practical approach to real‑time, privacy‑aware emotion recognition in automotive settings.
\end{abstract}


\section{Introduction}
Recent advances in machine learning (ML) have catalyzed innovations across multiple industries, including the automotive sector. An area of relevance is driver monitoring, where Emotion Recognition (ER) systems are emerging as enablers of enhanced Human-Machine Interaction (HMI) and personalized vehicle environments~\cite{fraunhofer2023humanmachine, Guel24_2}. By recognizing the emotional states of drivers in real time, these systems can promote safety and comfort by adapting the driving experience, for example, by adjusting seat position, regulating climate control, or activating stress-relief functions based on the driver’s mood. This personalization can enhance safety by addressing signs of fatigue, stress, or distraction~\cite{Guel24}.

Despite these promising applications, real-time ER in dynamic, uncontrolled environments like vehicle cabins raises practical concerns. Key obstacles include: (1) environmental variability, such as lighting changes, head poses, and occlusions (e.g., sunglasses), which complicate visual emotion detection systems~\cite{xu2021facial}; (2) physiological heterogeneity, where individual differences in heart rate, electrodermal activity (EDA), and other bodily responses to emotion are not well captured by single-modality systems; and (3) privacy concerns, as centralized models require continuous transmission of sensitive data like facial images and physiological signals, risking user privacy~\cite{sharma2020privacy}. Furthermore, traditional centralized approaches face scalability and computational inefficiencies in handling large amounts of real-time data from a diverse set of users.

In response to these challenges, we propose~\textbf{FedMultiEmo}, a Federated Learning (FL) framework for multimodal emotion recognition that preserves user privacy and addresses dynamic variability in real-time environments. FL, as a general paradigm for decentralized learning, allows model training to occur directly on user devices. This ensures that sensitive data, such as physiological signals or facial images, remains local, thereby significantly mitigating privacy risks~\cite{li2021federated}. Furthermore, FL distributes the computational load across edge devices, alleviating the pressure on central servers and allowing for scalable, real-time model updates~\cite{nakip2023decentralized}. It is particularly well-suited for automotive applications~\cite{Guel_2023}, where personalized comfort must be achieved in a privacy-conscious manner, and computational efficiency is essential.

Recent works have explored the integration of visual (e.g., facial expressions) and physiological (e.g., heart rate, EDA) modalities in emotion recognition~\cite{mollahosseini2019affectnet, ghosh2019multimodal}. However, common approaches either focus on a single modality or rely on centralized architectures that do not scale well to real-time applications. In contrast, multimodal FL enables real-time emotion detection that accounts for both visual and physiological cues. While FL has been successfully applied in privacy-preserving domains such as healthcare~\cite{sharma2024fedphy}, its integration into automotive emotion recognition systems is still emerging.

In this paper, we present a multimodal, real-time emotion recognition system for automotive environments using FL. Specifically, we integrate visual data (facial expressions) processed through a Convolutional Neural Network (CNN) with physiological data (heart rate, EDA, and skin temperature) classified via a Random Forest (RF) model. Our key contributions include: (1)~the design and implementation of a multimodal FL pipeline for emotion recognition, (2)~an edge-to-cloud deployment architecture using Raspberry Pi clients and a Flower server for decentralized learning, and (3)~a privacy-preserving, personalized aggregation mechanism based on Federated Averaging (FedAvg), weighted by client dataset sizes.

Experiments conducted on publicly available and self-collected datasets highlight the practical applicability of~\textbf{FedMultiEmo}, achieving 77\% accuracy for the visual CNN model, 74\% for the RF model on physiological data, and 87\% with multimodal decision-level fusion. We further show that our federated system converges in 18–20 rounds with an average round time of 120 seconds, offering real-time, privacy-conscious emotion recognition while maintaining an efficient computational footprint on client devices.

The remainder of this paper is structured as follows: Section~\ref{related_work} reviews related work on emotion recognition, multimodal fusion, and federated learning. Section~\ref{concept} outlines the design of the proposed multimodal federated learning framework. Section~\ref{prototype_results} describes the system setup and experiments and presents the results. Finally, Section~\ref{conclusion} concludes the paper and discusses future research directions.

\section{Related Work}
\label{related_work}

In this section, we review the state-of-the-art in automotive ER, multimodal learning strategies, and the role of FL in privacy-preserving emotion detection. We highlight the limitations of existing systems and position our contributions within this evolving landscape.

\subsection{Emotion Recognition in Automotive Contexts}

In-vehicle emotion recognition has become a focal point in efforts to enhance personalization and safety in modern vehicles. Commercial solutions such as Affectiva Automotive AI~\cite{affectiva} and Seeing Machines' Driver Monitoring Systems~\cite{tawari2021seeing} rely primarily on visual and audio cues for real-time monitoring of driver emotions. Fraunhofer IIS's multimodal systems combine biosignal monitoring with computer vision for stress and fatigue detection~\cite{iis-fraunhofer}. While effective in controlled settings, these systems often falter in real-world conditions characterized by varying lighting, occlusions (e.g., sunglasses), and head poses~\cite{zhao2020realworld}.

Critically, most commercial systems underutilize physiological signals, such as heart rate variability (HRV) and electrodermal activity (EDA), which offer objective, less environmentally-dependent indicators of emotional state~\cite{tripathi2021physio}. Furthermore, centralized processing models employed in commercial ER systems raise significant privacy concerns, as they require continuous transmission of sensitive personal data~\cite{ojs-aaai}.

Our proposed framework, FedMultiEmo, advances these commercial efforts by integrating both facial expression data and physiological signals while preserving privacy. This ensures robust performance in real-world conditions and eliminates the need for transmitting raw data to central servers.

\subsection{Multimodal Emotion Recognition}

Multimodal ER systems that combine visual, auditory, and physiological data outperform unimodal models by offering a holistic understanding of affective states~\cite{verma2022multimodal}. Early work by~\cite{healey2019, zeng2019} used physiological data with traditional classifiers such as Support Vector Machines (SVMs) and Random Forests (RFs), but their effectiveness was limited in real-time applications and noisy environments like car cabins.

Recent advances leverage deep learning (DL) models such as CNNs and Recurrent Neural Networks (RNNs) to extract meaningful patterns across modalities~\cite{ghosh2020, gupta2021}. These models improve classification accuracy but are computationally intensive and often unsuitable for edge deployment without significant optimization. Furthermore, current multimodal approaches struggle with data synchronization, modality-specific noise, and user variability in physiological responses~\cite{cui2025physiosync, udahemuka2024multimodal}.

Our work overcomes these limitations by introducing an edge-deployable multimodal ER system with decision-level fusion. By leveraging CNNs for visual input and RFs for physiological data, we balance model accuracy with computational efficiency. Through FL, we enable personalized model training across distributed devices, addressing variability and enhancing system adaptability.

\subsection{Federated Learning for Emotion Detection}

FL enables model training across user devices without transferring raw data, making it ideal for sensitive applications such as healthcare and driver monitoring~\cite{mcmahan2017communication, li2021federated}. In ER, FL has demonstrated promising results in physiological and EEG-based systems. For example, Fed-PhyERS~\cite{sharma2024fedphyers} processes physiological signals for privacy-aware emotion detection, while Reference~\cite{zhou2022federated} applied FL to EEG datasets, achieving over 90\% accuracy on SEED and DEAP datasets. FedCMD~\cite{li2024fedcmd} extended FL to cross-modal scenarios, showcasing the potential of decentralized training for multimodal affective computing. However, FL-based ER systems have not been widely applied to the automotive domain, particularly in scenarios requiring real-time processing and heterogeneous modalities. Existing methods often do not support multimodal fusion on edge devices or personalization based on individual user data.

Our contribution fills this gap by presenting a fully-integrated, real-time FL system tailored for in-vehicle multimodal ER. Unlike prior works, we deploy FL across Raspberry Pi clients using Flower and introduce a decision-level fusion strategy that combines CNN- and RF-based classifiers. Our personalization mechanism leverages FedAvg with client-weighted updates, improving generalization across heterogeneous users while maintaining privacy.

Although prior work has made important strides in emotion recognition in centralized and federated settings, limitations remain, particularly for real-time, multimodal, and privacy-conscious learning in automotive contexts. Across the reviewed approaches, most do not support fusion of heterogeneous modalities on edge devices, nor do they provide mechanisms for user-level personalization. These gaps motivate our proposed framework, which integrates FL with lightweight multimodal processing and edge deployment.

\par\vspace{0.5\baselineskip}

In summary, our work contributes:
\begin{itemize}
\item A real-time multimodal emotion recognition system for automotive use, implemented using FL.
\item A privacy-preserving personalization scheme based on on-device model training and edge deployment.
\item A decision-level multimodal fusion approach validated under real-world conditions.
\end{itemize}
These contributions position our proposed framework as a step toward practical deployment and align with current research directions in intelligent vehicles and edge learning.

\section{FedMultiEmo: A Multimodal Federated Learning Framework for Real-Time Emotion Recognition}
\label{concept}

\begin{figure*}[t]
  \centering
  \includegraphics[scale=1]{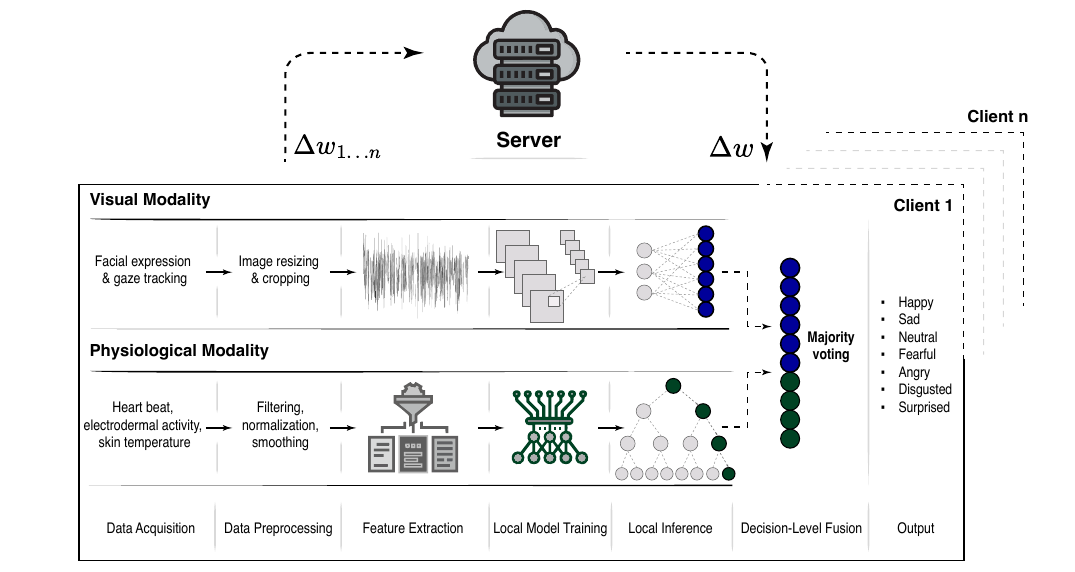}
  \caption{Pipeline for FedMultiEmo: A Multimodal Federated Learning Approach for Real-time Emotion Recognition, Illustrating the Key Stages.}
  \label{fig:multimodality_concept}
\end{figure*}

We introduce our proposed framework \textbf{FedMultiEmo}, a privacy‐preserving, multimodal federated learning framework designed for real‐time emotion recognition in vehicle cabins. It addresses three core challenges: (i) environmental variability in visual sensing, (ii) physiological heterogeneity across users, and (iii) privacy concerns inherent to centralized data collection. 
Fig.~\ref{fig:multimodality_concept} depicts the system pipeline, which consists of six stages, consistent across both visual and physiological modalities: \textit{Data Acquisition, Data Preprocessing, Feature Extraction, Local Model Training, Local Inference, Decision‐Level Fusion}. Furthermore, a \textit{Federated Aggregation} step synchronizes model updates across distributed clients.
%Fig.~\ref{fig:multimodality_concept} depicts the system pipeline, which consists of six stages: \textit{Data Preprocessing, Feature Extraction, Local Model Training, Local Inference, Decision‐Level Fusion, and Federated Aggregation}.

\subsection{Problem Formulation and Data Acquisition}
Let each client \(n \in \{1, \dots, N\}\) passively collect synchronized facial images \(\{I_t\}_{t \in \mathcal{T}}\) and physiological signals \(\{s_t\}_{t \in \mathcal{T}}\) during a driving session over a time window \(\mathcal{T}\). The goal is to collaboratively learn a global classifier \(f \colon (I, s) \mapsto y\) for predicting emotion labels \(y \in \mathcal{C}\), while preserving user privacy by keeping raw data local.

Each client maintains a private dataset \(D_n=\{(I_i, s_i, y_i)\}_{i=1}^{m_n}\), and the federated training objective is defined as:

\begin{align}
\label{eq:global_loss}
\mathcal{L}_{\mathrm{global}}(w) 
&= \sum_{n=1}^N \frac{m_n}{M}\, \mathcal{L}_n(w), \\
\mathcal{L}_n(w) 
&= \frac{1}{m_n} \sum_{(I, s, y) \in D_n} \ell(f_w(I, s), y),
\end{align}
where \(M = \sum_{n=1}^N m_n\) is the total number of local samples, and \(\ell\) is the cross-entropy loss used for training the visual CNN component. The physiological classifier, implemented as a Random Forest, uses Gini impurity during local training and does not participate in gradient-based optimization.

\subsection{Data Preprocessing}
We outline the preprocessing steps for the visual and physiological modalities used in the proposed emotion recognition framework.

\paragraph{Visual Modality}  
Facial frames are captured by the onboard camera at a resolution of \(640 \times 480\) pixels and resized to a standard \(48 \times 48\) grayscale format. The transformation from the original frame \(I(x, y)\) to the resized image \(I'(x', y')\) is performed using the following equation:
\begin{equation}
I'(x', y') = I\left(\left\lfloor x' \frac{h}{48} \right\rfloor, \left\lfloor y' \frac{w}{48} \right\rfloor \right),
\end{equation}
where \(I(x, y)\) represents the original image with height \(h~=~640\) and width \(w~=~480\), and \(I'(x', y')\) is the resized output. The indices \(x'\) and \(y'\) correspond to the pixel positions in the resized image. During training, image augmentation techniques such as horizontal flips and random rotations are applied to increase the robustness of the model and to account for variations in the orientation of the driver.

\paragraph{Physiological Modality}  
Physiological signals such as heart rate, electrodermal activity (EDA), and skin temperature are sampled at a frequency of 1 Hz, with each signal denoted as \(s(t)\), where \(t\) represents the time index in seconds. The preprocessing of these signals involves three key steps: filtering, smoothing, and normalization.

First, a 4th-order low-pass Butterworth filter with a cutoff frequency of 0.5 Hz is applied to each signal \(s(t)\) to remove high-frequency noise:
\begin{equation}
\tilde{s}[t] = \sum_{\tau=0}^{T} h[\tau] \cdot s[t - \tau],
\end{equation}
where \(h[\tau]\) represents the filter coefficients, and \(T\) is the length of the filter (4th-order filter, so \(T = 4\)).

Next, the filtered signal \(\tilde{s}(t)\) is smoothed using a 5-point moving average to reduce short-term fluctuations:
\begin{equation}
\hat{s}[t] = \frac{1}{5} \sum_{i=0}^{4} s[t - i],
\end{equation}
where \(\hat{s}[t]\) represents the smoothed version of the signal at time \(t\).

Finally, the smoothed signal \(\hat{s}[t]\) is normalized using z-score normalization, transforming the signal to have zero mean and unit variance:
\begin{equation}
\bar{s}[t] = \frac{\hat{s}[t] - \mu_{\hat{s}}}{\sigma_{\hat{s}}},
\end{equation}
where \(\mu_{\hat{s}}\) and \(\sigma_{\hat{s}}\) are the mean and standard deviation of the smoothed signal \(\hat{s}[t]\), respectively. The resulting \(\bar{s}[t]\) is the normalized signal used for feature extraction.

These preprocessing steps ensure that the physiological signals are clean, stable, and standardized, improving performance.

\subsection{Feature Extraction}
\label{sec:feature_extraction}

In this stage, both the visual and physiological features are extracted from the preprocessed data and prepared for classification. Feature extraction involves specialized models for each modality to capture the most discriminative information the recognition.

\paragraph{Visual Features}  
%The visual features are extracted using a CNN. The CNN architecture consists of three convolutional blocks, each followed by a ReLU activation and a max-pooling layer to reduce the spatial dimensions. The feature extraction process for the \(l\)-th layer can be mathematically expressed as:
Visual features are extracted using a CNN composed of three convolutional blocks. Each block includes a convolutional layer, followed by a ReLU activation and a max-pooling layer to reduce spatial dimensions. The feature extraction at the 
\(l\)-th  layer is formally defined as:
\begin{equation}
F_l = \mathrm{ReLU}(W_l * F_{l-1} + b_l), \quad F_0 = I',
\end{equation}
where \(W_l\) are the learnable convolutional filters at layer \(l\), \(b_l\) is the bias, and \(F_{l-1}\) is the input to the layer (which is the image or the feature map from the previous layer). The final output after passing through all layers is flattened and transformed into a fixed-length vector \(\mathbf{f}_{\mathrm{visual}} \in \mathbb{R}^{1024}\), which represents the visual features of the image. This feature vector is then passed to a softmax classifier to predict the emotional state over 7 possible classes.

\paragraph{Physiological Features}  
For physiological data, such as heart rate, EDA, and skin temperature, a set of handcrafted features is extracted from the normalized signal \(\bar{s}[t]\). For each physiological signal, the following features are computed over a 5-second window:

- Heart Rate Variability (HRV):  
  HRV is calculated as the mean absolute difference between consecutive heart rate measurements, capturing the variation in heart rate:
  \begin{equation}
  \mathrm{HRV} = \frac{1}{T} \sum_{i=1}^{T} \left| \bar{s}_{\mathrm{HR}}[i] - \bar{s}_{\mathrm{HR}}[i-1] \right|,
  \end{equation}
  where \(T\) is the number of samples in the window, and \(\bar{s}_{\mathrm{HR}}[i]\) represents the heart rate at the \(i\)-th sample.

- Maximum Electrodermal Activity (EDA):  
  The maximum value of the normalized EDA signal in the window is computed, providing an indicator of peak arousal:
  \begin{equation}
  \mathrm{EDA}_{\max} = \max_{i \in [T]} \bar{s}_{\mathrm{EDA}}[i],
  \end{equation}

- Temperature Fluctuations (\(\Delta T\)): 
  The temperature fluctuations are measured by the difference between the maximum and minimum skin temperature in the window, which offers insights into the autonomic nervous system's response to emotional stimuli:
  \begin{equation}
  \Delta T = \max \bar{s}_{\mathrm{Temp}} - \min \bar{s}_{\mathrm{Temp}}.
  \end{equation}
  
These three extracted features form the physiological feature vector \(\mathbf{f}_{\mathrm{physio}} \in \mathbb{R}^3\), which is then used as input to the Random Forest model for emotion classification. These features are then concatenated into a vector \(\mathbf{f}_{\mathrm{physio}} \in \mathbb{R}^3\) that serves as the input to the Random Forest classifier.

\subsection{Local Model Training}
\label{sec:local_training}

In the local model training phase, each client downloads the global model weights \(w^t\) from the server and performs local updates using its own dataset. Each client performs \(E = 4\) epochs of training, where the model is optimized using the Adam optimizer with a learning rate \(\eta = 10^{-4}\). Specifically, the update rule for the weights at client \(n\) is:

\begin{equation}
w_n^{t+1} = w^t - \eta \nabla \mathcal{L}_n(w^t),
\end{equation}
where \(w_n^{t+1}\) is the updated weight at client \(n\) after round \(t+1\), and \(\mathcal{L}_n(w^t)\) is the loss function for client \(n\) (cross-entropy loss for the visual CNN and Gini impurity for the Random Forest). Each client independently optimizes its model using its local dataset and computes updates for both the CNN and RF models.

\subsection{Local Inference}
\label{sec:local_inference}

Once local models are trained, each client makes predictions for both the visual and physiological modalities based on their respective classifiers. The final predictions are obtained by passing the input through the trained models.

\paragraph{Visual Prediction}  
For the visual modality, the input image \(I'\) is passed through the CNN, and a softmax operation is applied to obtain the predicted probabilities for each class:
\begin{equation}
\mathbf{p}_{\mathrm{visual}} = \mathrm{softmax}\left(\mathrm{CNN}_{w_{n,\mathrm{vis}}}(I')\right),
\end{equation}
where \(\mathbf{p}_{\mathrm{visual}} \in \mathbb{R}^7\) is the vector of predicted probabilities for the 7 emotion classes. The predicted emotion label is the class with the highest probability:
\begin{equation}
\hat{y}_{\mathrm{visual}} = \arg\max_k p_{\mathrm{visual},k},
\end{equation}
where \(p_{\mathrm{visual},k}\) represents the predicted probability of class \(k\).

\paragraph{Physiological Prediction}  
For the physiological modality, the extracted feature vector \(\mathbf{f}_{\mathrm{physio}}\) is passed through the Random Forest model, which outputs a probability distribution over the 7 emotion classes:
\begin{equation}
\mathbf{p}_{\mathrm{physio}} = \mathrm{RF}_{w_{n,\mathrm{phy}}}(\mathbf{f}_{\mathrm{physio}}).
\end{equation}
The predicted emotion label is again the class with the highest probability:
\begin{equation}
\hat{y}_{\mathrm{physio}} = \arg\max_k p_{\mathrm{physio},k}.
\end{equation}

\subsection{Decision-Level Fusion}
\label{sec:decision_fusion}

The final prediction is obtained through decision-level fusion of the predictions from both modalities. A majority voting mechanism is applied to select the emotion class with the highest aggregated probability across both the visual and physiological modalities. The final prediction is computed as:

\begin{equation}
\hat{y}_{\mathrm{final}} = \arg\max_{c \in \mathcal{C}} \left[ \mathbbm{1}(\hat{y}_{\mathrm{visual}} = c) + \mathbbm{1}(\hat{y}_{\mathrm{physio}} = c) \right],
\end{equation}
where \(\mathbbm{1}(\hat{y}_i = c)\) is an indicator function that returns 1 if the predicted class \(\hat{y}_i\) from modality \(i\) matches class \(c\), and 0 otherwise. This aggregation method reduces the impact of modality-specific noise or occlusions, ensuring a more robust and reliable prediction.

\subsection{Federated Aggregation}
\label{sec:fed_aggregation}

The Federated Averaging (FedAvg) algorithm is used to aggregate the local model updates from each client and produce a global model. At each round \(t+1\), the server aggregates the weight updates from all clients \(n = 1, \dots, N\) using the following equation:

\begin{equation}
w^{t+1} = \sum_{n=1}^N \frac{m_n}{M} w_n^{t+1},
\end{equation}
where \(w^{t+1}\) is the global model weight after round \(t+1\), \(w_n^{t+1}\) is the local model update at client \(n\), and \(m_n\) represents the number of data samples at client \(n\). The global model is updated as a weighted average of the client models, with each client's update weighted by the size of its local dataset. This aggregation strategy allows for personalization and ensures that the global model adapts to the data distributions at each client while preserving data privacy.

\vspace{1ex}
FedMultiEmo thus enables privacy-preserving, real-time emotion recognition by leveraging lightweight edge models, decision fusion, and federated optimization across multiple sensing modalities.

\begin{figure*}[t]  
    \centering
    \begin{subfigure}[b]{0.32\textwidth}
        \includegraphics[width=\textwidth]{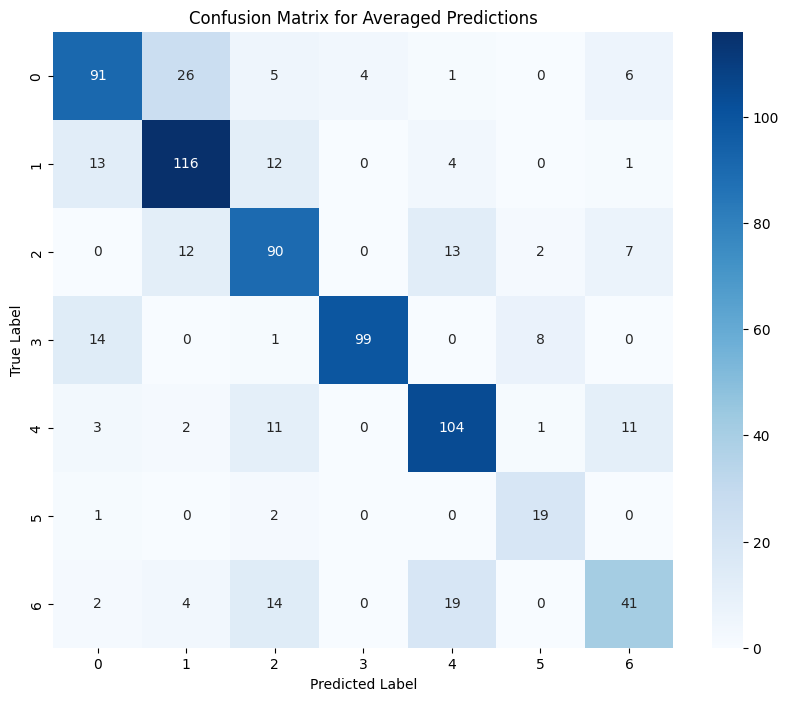}  
        \caption{Confusion Matrix of Physiological Model}
        \label{fig:fig1}
    \end{subfigure}
    \hfill
    \begin{subfigure}[b]{0.32\textwidth}
        \includegraphics[width=\textwidth]{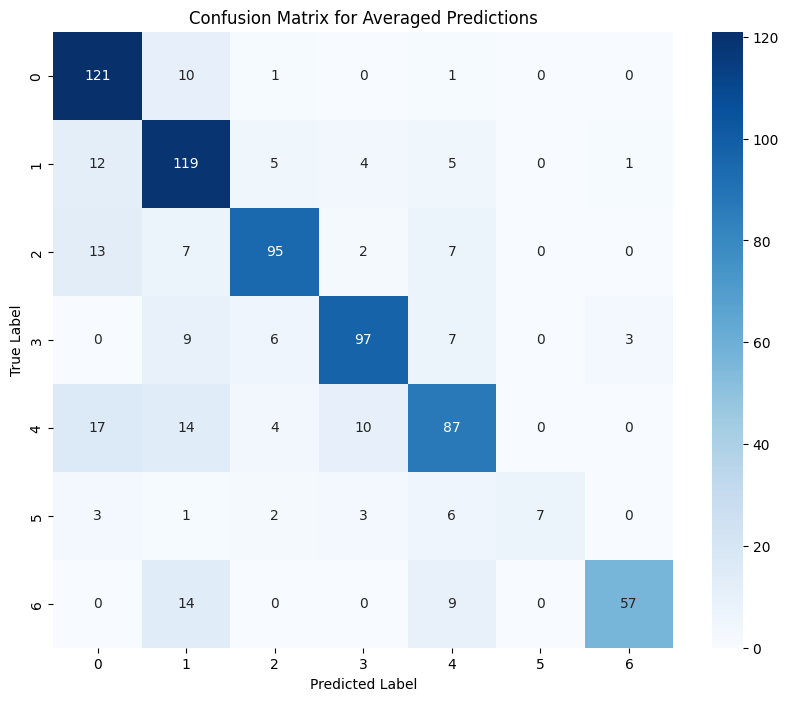}  
        \caption{Confusion Matrix of Image Model}
        \label{fig:fig2}
    \end{subfigure}
    \hfill
    \begin{subfigure}[b]{0.32\textwidth}
        \includegraphics[width=\textwidth]{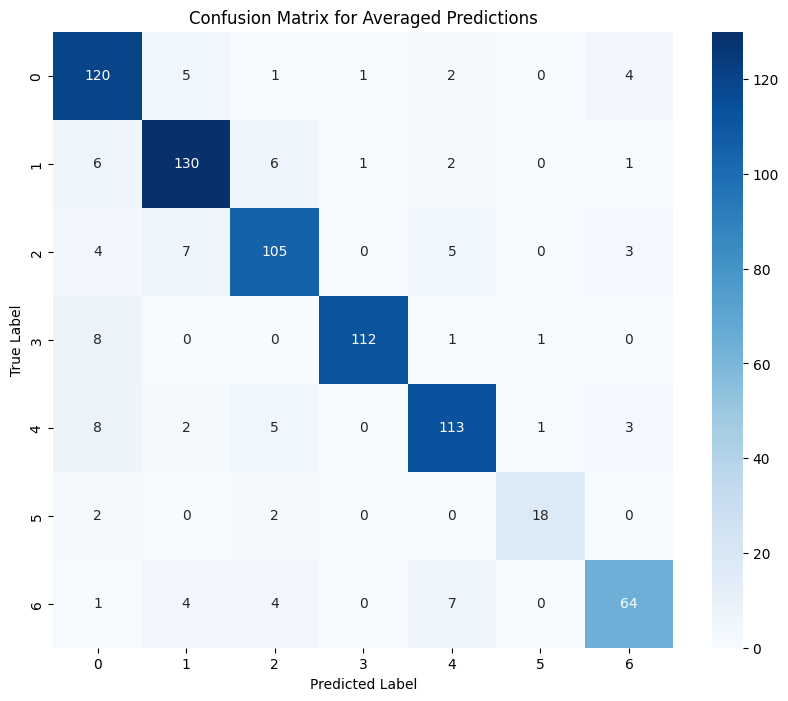}  
        \caption{Confusion Matrix of Multimodal Model}
        \label{fig:fig3}
    \end{subfigure}
    \caption{Comparison of Confusion Matrices for Physiological, Image, and Multimodal Data in 7-Emotion Classification.}
    \label{fig:confusion_matrices}
\end{figure*}

\section{Prototype Implementation and Experimental Results}
\label{prototype_results}

We describe the end‐to‐end implementation of our in‐vehicle, multimodal emotion recognition prototype and report its performance under practical FL settings. We first outline the two deployment phases and the hardware setup, then detail data collection and model architectures, finally presenting the FL configuration and resulting accuracy, convergence, and efficiency metrics.

\subsection{System Architecture and Hardware}
\label{sec:system_arch}

To validate both functionality and real‐time performance, we developed the prototype in two phases. In Phase I, rapid prototyping was performed on a single high‐performance laptop hosting the Flower federated server~\cite{beutel2020flower} and three Docker‐isolated client instances. Each client ingested prerecorded facial video and physiological traces (heart rate, EDA, and skin temperature) from local storage, enabling us to debug the end‐to‐end pipeline and FedAvg aggregation without hardware constraints.

Phase II transitioned to a true edge deployment. Three Raspberry Pi 4 devices served as clients, each paired with a USB camera and a Fitbit Versa 3 (via BLE) plus an I\(^2\)C‐connected EDA sensor for skin conductance. On each Pi, a TensorFlow client performed real‐time image capture, local CNN training/inference, physiological feature extraction, and Random Forest training. Only model weight updates traversed the WiFi link to the central laptop running Flower.

\subsection{Data Collection and Preprocessing}
\label{sec:data_preproc}
We combined a public vision dataset with a controlled, self‐collected physiological set to ensure robustness and privacy:

\paragraph{Visual Dataset}
We trained the facial expression model on FER2013~\cite{GOODFELLOW201559}, consisting of 35,887 \(48\times48\) grayscale images over seven emotion labels. To strengthen label quality, we adopted the FERPlus relabeling\cite{ferplus}: each image was annotated by ten crowdworkers, and consensus labels replaced the original ones. Images were rescaled by \(1/255\), face‐cropped to \(48\times48\), and augmented with random horizontal flips and \(\pm10^\circ\) rotations.

\paragraph{Physiological Dataset}
To avoid indirect valence–arousal proxies, we collected a bespoke dataset in our lab. Subjects wore a Fitbit Versa 3 and an I\(^2\)C EDA sensor while exposed to emotion‐eliciting video clips. Raw heart rate, HRV, and skin conductance streams were low‐pass filtered (4th‐order Butterworth, 0.5 Hz cutoff), z‑score normalized, and smoothed via a 5‑sample moving average. Ground truth emotion labels were synchronized to the video timeline.

\subsection{Model Architectures}
\label{sec:models}

Two independent classifiers process each modality:

\paragraph{Facial Expression CNN}
The CNN comprises three convolutional blocks—\((\mathrm{Conv}+ \mathrm{ReLU})\rightarrow\mathrm{MaxPool}\)—with filter depths \(\{32, 64, 128\}\), each followed by dropout~(\(p=0.25\)). Flattened features feed a 1024‐unit dense layer (ReLU, \(p=0.5\)), then a softmax output over seven emotions. The model is optimized with Adam (initial LR \(10^{-4}\), exponential decay).

\paragraph{Physiological Random Forest}
Physiological features, notably HRV, EDA peak rate, and temperature fluctuation over 5-second windows are classified by a 200‐tree Random Forest (Gini criterion). This nonparametric ensemble is robust to noisy signals and small sample sizes, producing per‐class probability estimates.

\subsection{Federated Learning Setup}
\label{sec:fl_setup}
We orchestrate decentralized training via Flower with the standard FedAvg algorithm. Each of the three clients performs four local epochs per round on its own data; model weight deltas and validation metrics are returned to the server. The server aggregates local updates with FedAvg and we executed 20 global rounds in both phases.

\subsection{Results}
\label{sec:results}
Classification performance is evaluated for unimodal and multimodal inputs; training efficiency is compared across individual, centralized, and federated setups.

\subsubsection{Classification Analysis}
The evaluation covers both unimodal and multimodal classification performance across three learning configurations: (a) physiological signals, (b) facial images, and (c) multimodal fusion, as illustrated in Fig.~\ref{fig:confusion_matrices}.

Fig.~\ref{fig:confusion_matrices}(a) shows the confusion matrix for the Random Forest classifier trained solely on physiological features. The model achieves a test accuracy of approximately 74\%. Performance across emotion classes is uneven, primarily due to class imbalance. The \textit{disgust} class (class 5) exhibits the lowest true positive rate, reflecting its limited representation in the dataset. Overall, the physiological model achieves precision and recall of 67\% and 73\%, respectively.

Fig.~\ref{fig:confusion_matrices}(b) presents the results for facial image classification using a CNN. This model achieves a higher test accuracy of 78\% and demonstrates more balanced performance across classes. True positive rates increase across most emotions, although \textit{disgust} remains challenging due to limited training samples and its subtle facial cues. Still, the visual modality provides stronger generalization, with both precision and recall averaging around 77\%, outperforming the physiological-only approach.

Fig.~\ref{fig:confusion_matrices}(c) displays the confusion matrix for the decision-level fusion model, which combines outputs from both modalities. This configuration achieves the highest overall accuracy of 87\%. Notably, the \textit{disgust} class shows a marked improvement in true positive rate, illustrating the complementary nature of multimodal input. Precision, recall, and F1-scores average 87\% across all classes. These findings confirm that multimodal fusion not only mitigates individual modality limitations but also enhances robustness in real-world conditions such as poor lighting or sensor noise. %Furthermore, the framework's implementation on Raspberry Pi devices demonstrates its practical viability for in-vehicle, privacy-preserving emotion recognition.

\subsubsection{Training Efficiency and Time Analysis}
In addition to classification analysis, we compare the training efficiency across three setups: individual, centralized, and federated learning. Table~\ref{tab:training_time} summarizes training times, client distribution, and configuration details.

\begin{table}[h]
\centering
\caption{Training Time Comparison Across Learning Paradigms.}
\label{tab:training_time}
\begin{tabular}{lccc}
\toprule
\textbf{Parameter} & \textbf{Individual} & \textbf{Centralized} & \textbf{Federated} \\
\midrule
Time Taken (s) & 430 & 1350 & 4300 \\
No. of Clients & 1 & 3 & 3 \\
Epochs / Rounds & 4 & 40 & 10 rounds × 4 epochs \\
Time / Round (s) & – & – & 430 \\
Time / Epoch (s) & 110 & 42 & 110 (local/client) \\
\bottomrule
\end{tabular}
\end{table}

%In the individual learning setup, a single client trains for four epochs, requiring 430 seconds. Centralized learning aggregates data from three clients and trains for 40 epochs, completing in 1350 seconds, with an average of 42 seconds per epoch.

%FL involves 10 communication rounds, each consisting of 4 local epochs per client. This setup takes the longest, approximately 4300 seconds in total, with each round lasting about 430 seconds, driven by local computation (110 seconds per epoch per client) and communication overhead. While FL incurs a higher time cost, it offers the advantage of preserving data locality and user privacy. This trade-off is appropriate in scenarios with strong privacy requirements, such as in-vehicle driver monitoring.

In the individual learning setup, a single client trains for four epochs, requiring 430 seconds. Centralized learning aggregates data from three clients and trains for 40 epochs, completing in 1350 seconds, with an average of 42 seconds per epoch. FL involves 10 communication rounds, each consisting of 4 local epochs per client. This setup takes the longest, approximately 4300 seconds in total, with each round lasting about 430 seconds, driven by local computation (110 seconds per epoch per client) and communication overhead. All timings are relative to the time base maintained by the FL server, which is periodically disciplined via Network Time Protocol (NTP) to limit drift and provide a consistent temporal baseline for the reported measurements~\cite{tz2024,syncfed,10092630}. While FL incurs a higher time cost, it offers the advantage of preserving data locality and user privacy. This trade-off is appropriate in scenarios with strong privacy requirements, such as in-vehicle driver monitoring.

\section{Conclusion and Future Work}
\label{conclusion}
In this paper, we introduced \textbf{FedMultiEmo}, a privacy-preserving multimodal federated learning framework for real-time emotion recognition in vehicle cabins. By integrating facial expressions analyzed through Convolutional Neural Networks with physiological signals processed via Random Forest classifiers, the framework addresses challenges related to \textit{environmental variability}, \textit{physiological diversity}, and \textit{data privacy}. A decision-level fusion mechanism balances these modalities, while the federated averaging approach supports scalable, personalized learning across clients without sharing raw data.

Implemented on Raspberry Pi edge devices and a Flower server, \textbf{FedMultiEmo} performs consistently, with the multimodal classifier attaining 87\% accuracy, surpassing centralized baselines while maintaining data locality and adhering to strict resource constraints. The system converges within 18 rounds, while keeping communication and memory requirements within practical limits. The primary contributions of this work are:
\begin{itemize}
  \item \emph{Multimodal Fusion in FL:} An in‑vehicle system to fuse CNN and RF predictions via decision‐level voting in a federated setup.
  \item \emph{Edge‐to‐Cloud Deployment:} A working prototype on Raspberry Pi clients and a Flower server, showing feasibility under real-time constraints.
  \item \emph{Privacy‐Preserving Personalization:} A personalization scheme based on FedAvg, using client data volume to adapt global models without raw data exchange.
\end{itemize}

Looking ahead, we plan to advance \textbf{FedMultiEmo} in three key areas: (i) transitioning from hand-crafted physiological features to learned temporal embeddings using models like LSTMs or Transformers; (ii) developing adaptive, confidence-weighted fusion strategies to enhance decision-level voting; and (iii) addressing real-world deployment challenges, including missing modalities, asynchronous client updates, and diverse hardware capabilities. Finally, validating the framework on a wider range of real-world datasets will strengthen its generalizability.

\bibliographystyle{ieeetr}
\bibliography{references_emo}

\begin{thebibliography}{10}

\bibitem{fraunhofer2023humanmachine}
F.~I. for Integrated Circuits~IIS, ``Human-machine interaction and driver monitoring systems,'' {\em Fraunhofer IIS}, 2023.
\newblock Accessed: 2025-03-03.

\bibitem{Guel24_2}
B.~C. Gül, D.~Dittler, N.~Jazdi, and M.~Weyrich, ``Federated learning for comfort features in vehicles with collaborative sensing: A review,'' in {\em 2024 IEEE 29th International Conference on Emerging Technologies and Factory Automation (ETFA)}, pp.~1--7, 2024.

\bibitem{Guel24}
B.~C. Gül, N.~Devarakonda, N.~Jazdi, and M.~Weyrich, ``Personalized comfort features in software-defined vehicles using federated learning,'' in {\em 2024 IEEE 29th International Conference on Emerging Technologies and Factory Automation (ETFA)}, pp.~1--7, 2024.

\bibitem{xu2021facial}
Z.~Xu, J.~Li, and Y.~Wang, ``Challenges in facial emotion recognition under real-world variability,'' {\em IEEE Transactions on Affective Computing}, vol.~12, no.~1, pp.~52--64, 2021.

\bibitem{sharma2020privacy}
R.~Sharma, S.~Singh, and S.~Kaur, ``Privacy and security in emotion recognition systems: A review,'' {\em Journal of Privacy and Confidentiality}, vol.~8, no.~2, pp.~135--150, 2020.

\bibitem{li2021federated}
Q.~Li and X.~Yang, ``Federated learning for privacy-preserving emotion recognition from multimodal data,'' {\em IEEE Transactions on Neural Networks and Learning Systems}, vol.~32, no.~3, pp.~815--825, 2021.

\bibitem{nakip2023decentralized}
M.~Nakip, B.~C. G{\"u}l, and E.~Gelenbe, ``Decentralized online federated g-network learning for lightweight intrusion detection,'' in {\em 2023 31st international symposium on modeling, analysis, and simulation of computer and telecommunication systems (MASCOTS)}, pp.~1--8, IEEE, 2023.

\bibitem{Guel_2023}
B.~G{\"u}l, N.~Devarakonda, D.~Dittler, N.~Jazdi, and M.~Weyrich, ``Using federated learning in the context of software-defined mobility systems for predictive quality of service,'' vol.~2419, 2023.

\bibitem{mollahosseini2019affectnet}
A.~Mollahosseini, B.~Hasani, and M.~H. Mahoor, ``Affectnet: A database for facial expression, valence, and arousal computing,'' {\em IEEE Transactions on Affective Computing}, vol.~10, no.~1, pp.~18--31, 2019.

\bibitem{ghosh2019multimodal}
S.~Ghosh, P.~Sharma, and M.~Gupta, ``Multimodal emotion recognition using deep learning: A survey,'' {\em IEEE Access}, vol.~7, pp.~173643--173654, 2019.

\bibitem{sharma2024fedphy}
M.~Sharma, R.~Gupta, and S.~Agarwal, ``Fed-phyers: A federated learning approach for multimodal emotion recognition using physiological signals,'' {\em IEEE Access}, vol.~12, pp.~12564--12574, 2024.

\bibitem{affectiva}
Affectiva, ``Automotive ai.'' \url{https://www.affectiva.com}, 2023.
\newblock Accessed: 2023-03-03.

\bibitem{tawari2021seeing}
A.~Tawari, H.~Martin, and S.~Larsen, ``Seeing machines driver monitoring technology: A real-world application,'' {\em SAE International Journal of Connected and Automated Vehicles}, vol.~4, no.~1, pp.~23--38, 2021.

\bibitem{iis-fraunhofer}
{Fraunhofer Institute for Integrated Circuits IIS}, ``Active health monitoring solutions,'' 2023.
\newblock Accessed: 2023-03-03.

\bibitem{zhao2020realworld}
M.~Zhao, J.~Zhang, D.~Li, and J.~Sun, ``Real-world challenges in facial emotion recognition: A comprehensive survey,'' {\em IEEE Transactions on Affective Computing}, vol.~11, no.~2, pp.~380--394, 2020.

\bibitem{tripathi2021physio}
S.~Tripathi, A.~Agrawal, and P.~Verma, ``Physiological signal analysis for emotion recognition using random forest and feature selection,'' {\em International Journal of Biomedical Engineering and Technology}, vol.~37, no.~3, pp.~197--214, 2021.

\bibitem{ojs-aaai}
J.~Smith, K.~Li, and Y.~Zhang, ``Privacy and ethical concerns in in-vehicle emotion tracking technologies,'' {\em Proceedings of the AAAI Conference on Artificial Intelligence}, vol.~36, no.~1, pp.~4552--4561, 2022.

\bibitem{verma2022multimodal}
R.~Verma, N.~Kumar, and A.~Prakash, ``Multimodal fusion of physiological and visual cues for real-time driver emotion detection,'' {\em Neural Computing and Applications}, vol.~34, no.~5, pp.~12789--12805, 2022.

\bibitem{healey2019}
J.~Healey, S.~M. Sharma, and R.~Picard, ``Emotion recognition using physiological signals: A review,'' {\em IEEE Trans. Pattern Anal. Mach. Intell.}, vol.~41, no.~12, pp.~2961--2982, 2019.

\bibitem{zeng2019}
Z.~Zeng, J.~P. G.~L. Hu, and S.~J. Ho, ``Facial expression recognition: A literature review,'' {\em IEEE Trans. Affective Computing}, vol.~10, no.~3, pp.~313--327, 2019.

\bibitem{ghosh2020}
S.~Ghosh {\em et~al.}, ``End-to-end deep learning for emotion recognition in the wild,'' {\em IEEE Transactions on Affective Computing}, vol.~11, no.~4, pp.~558--570, 2020.

\bibitem{gupta2021}
P.~Gupta {\em et~al.}, ``Deep convolutional neural networks for emotion recognition: A survey,'' {\em IEEE Transactions on Neural Networks and Learning Systems}, vol.~32, no.~5, pp.~1312--1325, 2021.

\bibitem{cui2025physiosync}
K.~Cui, J.~Li, Y.~Liu, X.~Zhang, Z.~Hu, and M.~Wang, ``Physiosync: Temporal and cross-modal contrastive learning inspired by physiological synchronization for eeg-based emotion recognition,'' {\em arXiv preprint arXiv:2504.17163}, 2025.

\bibitem{udahemuka2024multimodal}
G.~Udahemuka, K.~Djouani, and A.~M. Kurien, ``Multimodal emotion recognition using visual, vocal and physiological signals: a review,'' {\em Applied Sciences}, vol.~14, no.~17, p.~8071, 2024.

\bibitem{mcmahan2017communication}
B.~McMahan, E.~Moore, D.~Ramage, S.~Hampson, and B.~A. y~Arcas, ``Communication-efficient learning of deep networks from decentralized data,'' in {\em Artificial intelligence and statistics}, pp.~1273--1282, PMLR, 2017.

\bibitem{sharma2024fedphyers}
R.~Sharma, A.~Gupta, and M.~Singh, ``Fed-phyers: Federated learning-based multi-modal emotion recognition using physiological signals,'' {\em Multimedia Tools and Applications}, 2024.

\bibitem{zhou2022federated}
M.~Zhou, X.~Liu, and Y.~Zhang, ``Federated learning for eeg-based emotion recognition,'' {\em Electronics}, vol.~11, no.~20, p.~3316, 2022.

\bibitem{li2024fedcmd}
J.~Li, X.~Wang, and H.~Xu, ``Fedcmd: Federated cross-modal distillation for driver emotion recognition,'' {\em Proceedings of the ACM on Interactive, Mobile, Wearable and Ubiquitous Technologies}, vol.~8, no.~1, pp.~1--23, 2024.

\bibitem{beutel2020flower}
D.~J. Beutel, T.~Topal, A.~Mathur, X.~Qiu, J.~Fernandez-Marques, Y.~Gao, L.~Sani, K.~H. Li, T.~Parcollet, P.~P.~B. de~Gusm{\~a}o, {\em et~al.}, ``Flower: A friendly federated learning research framework,'' {\em arXiv preprint arXiv:2007.14390}, 2020.

\bibitem{GOODFELLOW201559}
I.~J. Goodfellow, D.~Erhan, P.~{Luc Carrier}, A.~Courville, M.~Mirza, B.~Hamner, W.~Cukierski, Y.~Tang, D.~Thaler, D.-H. Lee, Y.~Zhou, C.~Ramaiah, F.~Feng, R.~Li, X.~Wang, D.~Athanasakis, J.~Shawe-Taylor, M.~Milakov, J.~Park, R.~Ionescu, M.~Popescu, C.~Grozea, J.~Bergstra, J.~Xie, L.~Romaszko, B.~Xu, Z.~Chuang, and Y.~Bengio, ``Challenges in representation learning: A report on three machine learning contests,'' {\em Neural Networks}, vol.~64, pp.~59--63, 2015.
\newblock Special Issue on “Deep Learning of Representations”.

\bibitem{ferplus}
E.~Barsoum, C.~Zhang, C.~C. Ferrer, and Z.~Zhang, ``Training deep networks for facial expression recognition with crowd-sourced label distribution,'' in {\em Proceedings of the 18th ACM international conference on multimodal interaction}, pp.~279--283, 2016.

\bibitem{tz2024}
S.~Tziampazis, P.~Hirmer, and M.~Weyrich, ``A hybrid framework for latency compensation in remote testing of automotive electronic control units,'' {\em Frontiers in the Internet of Things}, vol.~Volume 3 - 2024, 2024.

\bibitem{syncfed}
B.~C. Gül, S.~Tziampazis, N.~Jazdi, and M.~Weyrich, ``Syncfed: Time-aware federated learning through explicit timestamping and synchronization,'' 2025.

\bibitem{10092630}
S.~Tziampazis, O.~Kopp, and M.~Weyrich, ``Distributed integration of electronic control units for automotive oems: Challenges, vision, and research directions,'' in {\em 2023 IEEE 20th International Conference on Software Architecture Companion (ICSA-C)}, pp.~296--300, 2023.

\end{thebibliography}
\end{document}